\documentclass[10pt,twocolumn,letterpaper]{article}

\usepackage[pagenumbers]{cvpr} 
\usepackage[marginal]{footmisc}

\definecolor{cvprblue}{rgb}{0.21,0.49,0.74}
\usepackage[pagebackref,breaklinks,colorlinks,allcolors=cvprblue,urlcolor=magenta]{hyperref}
\usepackage{url}            
\usepackage{booktabs}       
\usepackage{amsfonts}       
\usepackage{nicefrac}       
\usepackage{microtype}      
\usepackage{xcolor}         
\usepackage{wrapfig}
\usepackage{ulem}
\usepackage{soul}

\usepackage{xcolor}
\usepackage{enumitem}
\usepackage{xspace}
\usepackage{booktabs}
\usepackage{colortbl}
\usepackage{multirow}
\usepackage{makecell}
\usepackage{lipsum} 
\usepackage{graphicx}
\usepackage{algorithmic}
\usepackage{algorithm}
\usepackage{titletoc}

\usepackage[labelsep=period]{caption}
\renewcommand{\paragraph}[1]{\vspace{0.2em}\noindent \textbf{#1 }}

\definecolor{MyDarkRed}{rgb}{0.66, 0.16, 0.16}
\definecolor{MyDarkBlue}{rgb}{0.16, 0.16, 0.66}

\definecolor{Red}{rgb}{0.6,0,0}
\definecolor{Blue}{rgb}{0,0,0.8}
\definecolor{Green}{rgb}{0.2,0.8,0}
\definecolor{airforceblue}{rgb}{0.36, 0.54, 0.66}
\definecolor{ao(english)}{rgb}{0.0, 0.5, 0.0}
\definecolor{azure(colorwheel)}{rgb}{0.0, 0.5, 1.0}
\definecolor{crimson}{rgb}{0.86, 0.08, 0.24}
\definecolor{darkcerulean}{rgb}{0.03, 0.27, 0.49}
\definecolor{cobalt}{rgb}{0.0, 0.28, 0.67}
\definecolor{rosegold}{rgb}{0.72, 0.43, 0.47}
\definecolor{orange-red}{rgb}{1.0, 0.27, 0.0}
\definecolor{mountainmeadow}{rgb}{0.19, 0.73, 0.56}
\definecolor{malachite}{rgb}{0.04, 0.85, 0.32}
\definecolor{darkblue}{rgb}{0.0, 0.0, 0.55}
\definecolor{customblue}{rgb}{0.2, 0.35, 0.8}

\newcommand{\latent}{\boldsymbol{z}}

\newcommand{\token}{\mathbf{T}}

\usepackage{amsmath,amsfonts,bm}

\def\eqref#1{equation~\ref{#1}}

\def\1{\bm{1}}

\DeclareMathAlphabet{\mathsfit}{\encodingdefault}{\sfdefault}{m}{sl}
\SetMathAlphabet{\mathsfit}{bold}{\encodingdefault}{\sfdefault}{bx}{n}

\def\confName{CVPR}
\def\confYear{2026}
\renewcommand{\thefootnote}{}
\newcommand{\method}{LumiTex}

\title{\method: Towards High-Fidelity PBR Texture Generation\\with Illumination Context}

\author{%
\hspace{-.7em}%
Jingzhi Bao$^{1}$\footnotemark[1]\;,\;
Hongze Chen$^{2}$\thanks{Equal contribution.}\;,\;
Lingting Zhu$^{3}$,\;
Chenyu Liu$^{4}$,\;
Runze Zhang$^{5}$,\;
Keyang Luo$^{5}$,\;
Zeyu Hu$^{5}$,\;\\[0.1em]
\hspace{-.7em}Weikai Chen$^{5}$,\;
Yingda Yin$^{5}$,\;
Xin Wang$^{5}$,\;
Zehong Lin$^{2}$\footnotemark[2],\;
Jun Zhang$^{2}$,\;
Xiaoguang Han$^{1}$\thanks{Corresponding author.} \\[0.2em]
{$^1$CUHK(SZ)}\quad{$^2$HKUST}\quad
{$^3$HKU}\quad{$^4$PKU}\quad{$^5$LIGHTSPEED}
}

\begin{document}

\twocolumn[{%
\renewcommand\twocolumn[1][]{#1}%
\maketitle
\vspace{-32pt}
\begin{center}
    \centering
    \captionsetup{type=figure}
    \includegraphics[width=\linewidth]{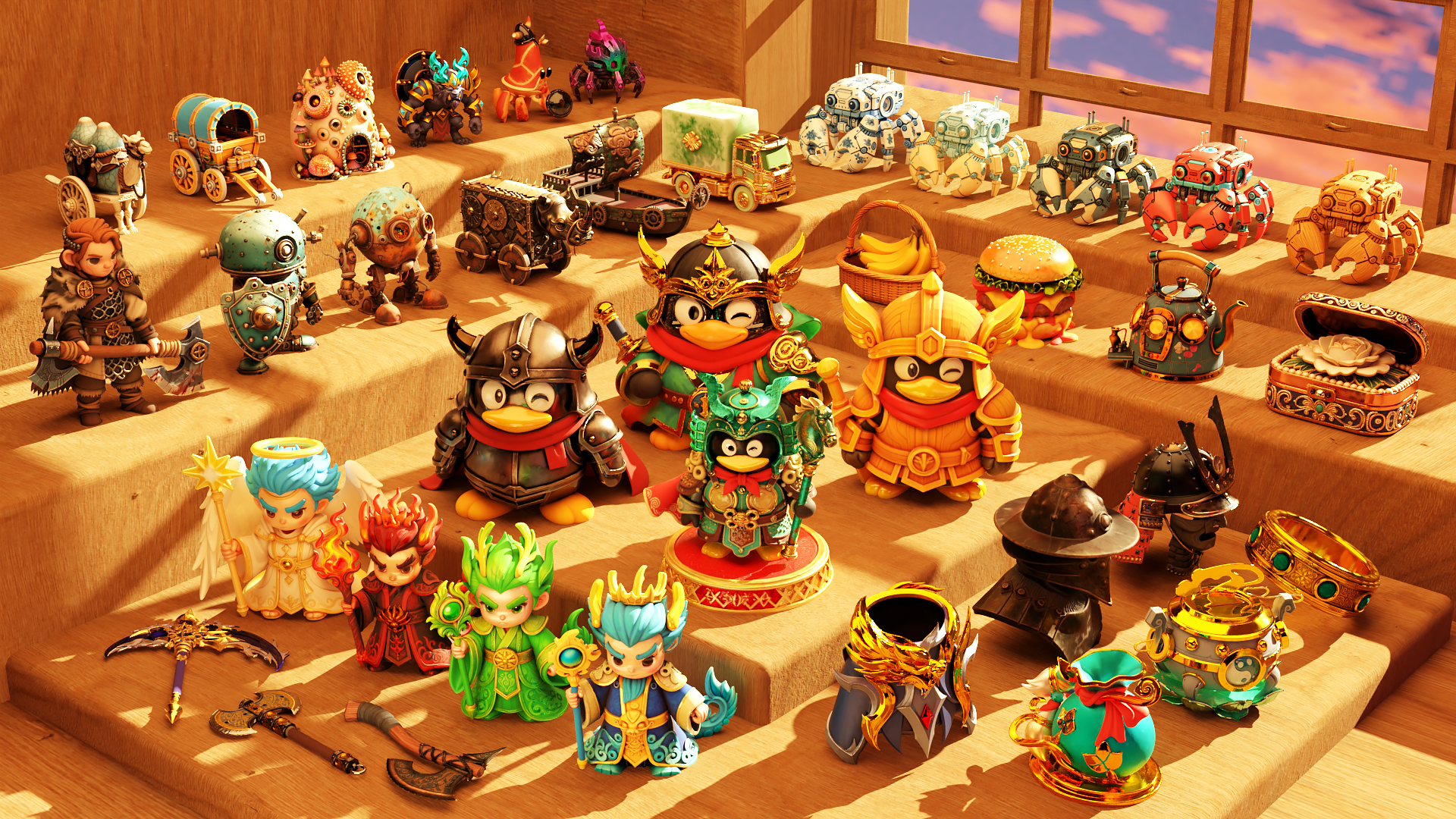}
    \vspace{-1.3em}
    \captionof{figure}{A collection of textured meshes with PBR materials generated by \method, capturing both intricate detail and convincing physical realism across diverse object categories under environmental lighting. Our demo is available at: \tt \href{https://lumitex.vercel.app}{https://lumitex.vercel.app}.}
    \label{fig:teaser}
\end{center}%
}]

\def\thefootnote{}\footnotetext{$*$ The first two authors contributed equally. $\dagger$ Corresponding authors.}

\newif\ifdrafting
\draftingtrue
\ifdrafting
    \newcommand{\jz}[1]{\textcolor{cyan}{[JZ: #1]}}
    \newcommand{\hz}[1]{\textcolor{customblue}{[HZ: #1]}}
\else
    \newcommand{\jz}[1]{}
    \newcommand{\hz}[1]{}
\fi

\begin{abstract}

Physically-based rendering (PBR) provides a principled standard for realistic material-lighting interactions in computer graphics. 
Despite recent advances in generating PBR textures, existing methods fail to address two fundamental challenges:
1) materials decomposition from image prompts under limited illumination cues, and
2) seamless and view-consistent texture completion.
To this end, we propose LumiTex, an end-to-end framework that comprises three key components:
(1) a multi-branch generation scheme that disentangles albedo and metallic-roughness under shared illumination priors for robust material understanding,
(2) a lighting-aware material attention mechanism that injects illumination context into the decoding process for physically grounded generation of albedo, metallic, and roughness maps,
and (3) a geometry-guided inpainting module based on a large view synthesis model that enriches texture coverage and ensures seamless, view-consistent UV completion.
Extensive experiments demonstrate that LumiTex achieves state-of-the-art performance in texture quality, surpassing both existing open-source and commercial methods.

\end{abstract}

\section{Introduction}
\label{sec:intro}

Physically-based rendering (PBR) is the industry standard for material and lighting representation in games, films, and AR/VR. 
PBR textures encode key surface properties such as albedo and metallic–roughness (MR), allowing for realistic visual interactions under diverse lighting conditions. Nevertheless, it is challenging to generate PBR textures, which requires both accurate physical characterization of materials and consistency across multiple viewpoints.

\begin{figure}[t]
    \centering
    \includegraphics[width=\linewidth]{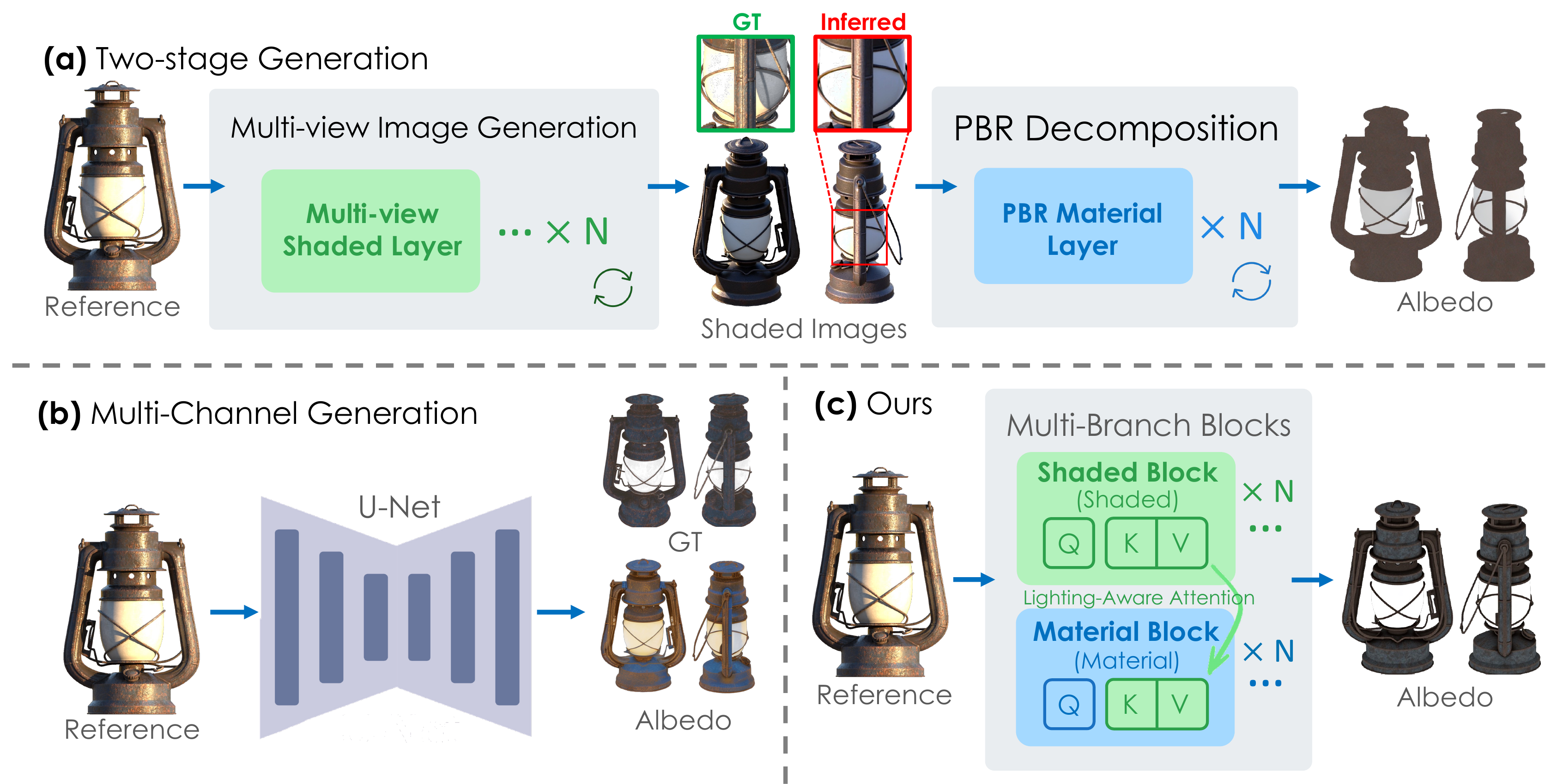}
    \caption{\textbf{Illustration of Different PBR Modeling.} 
    Unlike (a) two-stage PBR texture generation with suboptimal intermediates, 
    and (b) end-to-end approach w/o multi-view shaded features, 
    (c) our multi-branch design leverages multi-view consistent lighting features and surpasses the generation quality of other models.}
    \label{fig:pipe_design}
    \vspace{-1.8em}
\end{figure}

Recent advances in PBR texture generation, primarily driven by the remarkable capabilities of generative models, can be broadly categorized into two main approaches. The first line of research~\citep{zhang2024dreammat,zhu2025muma,hong2024supermat,munkberg2025videomat} adopts a two-stage approach: it first generates multi-view images that encode baked environment lighting (shaded images), and then obtains PBR textures either through optimization or by employing dedicated multi-view inference models, e.g., IDArb~\citep{li2025idarb}. These shaded images supply rich illumination cues that are crucial for accurate material decomposition. However, optimization-based methods such as DreamMat~\citep{zhang2024dreammat} are severely limited by long optimization time. While multi-view inference models improve efficiency, existing methods like MuMA~\citep{zhu2025muma} often produce inferior materials, as the generated intermediates are frequently suboptimal and poorly aligned with the training inputs, as illustrated in Fig.~\ref{fig:pipe_design}(a).

%
The second stream~\citep{zhao2025hunyuan3d,he2025materialmvp,zhu2024mcmat,materialanything} focuses on 
jointly generating multi-view albedo, metallic, and roughness images with multi-channel diffusion (see Fig.~\ref{fig:pipe_design}(b)) and subsequently baking multi-view images to the UV space via camera projection~\citep{chen2023text2tex,richardson2023texture}.
However, these methods suffer from three key limitations. 
Firstly, generating accurate multi-view material images is challenging since image diffusion models~\citep{ho2020denoising,podell2023sdxl,lin2024common,esser2024sd3} lack sufficient material priors, and reference images provide only limited illumination cues.
Secondly, the domain gap between albedo and metallic-roughness (MR) is often overlooked. Albedo reflects the intrinsic surface color, whereas MR encodes illumination-dependent material properties. However, multi-channel approaches typically predict them jointly within a shared output space.
This entangled representation, without explicit contextual guidance, hinders accurate material decomposition and compromises the physical plausibility of the generated textures.
Lastly, existing datasets suffer from severe data imbalance. While shaded and albedo images are relatively abundant, high-quality metallic and roughness maps are much scarcer, limiting effective supervision of PBR materials.

To address these limitations, we propose LumiTex, an end-to-end framework that jointly generates multi-view shaded images and PBR maps. Our key idea is to integrate \textit{multi-stage inference} into a single one to overcome the reliance on imperfect shaded intermediates while alleviating the data imbalance and retaining rich illumination cues via \textit{multi-stage training}.
Specifically, we first train a multi-view illumination context branch to reconstruct shaded images across views, capturing rich and consistent lighting cues as an explicit illumination context.
To address the domain gap between albedo and MR in joint modeling, we then introduce a lighting-aware material attention mechanism in the material branches that separately guide albedo and MR by the illumination context for channel-specific reasoning.
By combining the generative prior of diffusion models with the rich illumination context extended from reference images, \method~enables high-fidelity and physically plausible texture generation in a fully end-to-end manner.

To further enhance the quality and completeness of generated textures, we introduce a general texture inpainting strategy based on a large view synthesis model (LVSM) ~\citep{jin2025lvsm}.
Specifically, we train a geometry-guided LVSM to synthesize novel views for missing or occluded regions based on the generated views. 
Given input views, geometry maps, and camera poses, the model synthesizes new viewpoints from arbitrary target poses, effectively densifying the texture observations. In contrast to UV-based methods~\citep{yu2023texture, cheng2024mvpaint,yu2024texgen,zeng2024paint3d}, which suffer from discontinuities and topological ambiguities in UV space, our approach performs view-space completion, producing seamless and globally consistent textures.

The contributions of this work are summarized as follows:
\begin{itemize}
    \item We propose an end-to-end multi-branch framework for high-quality PBR texture generation, where a multi-view illumination context branch captures consistent lighting priors to alleviate data imbalance and reliance on imperfect intermediates in two-stage designs.
    \item We introduce a lighting-aware material attention mechanism that conditions albedo and metallic–roughness generation on shared illumination priors, enabling disentangled material reasoning and improving physical plausibility.
    \item We propose an advanced texture inpainting strategy, leveraging LVSM to extend the generated views to a denser set for seamless and globally consistent texture completion.
    \item Extensive experiments demonstrate that our method surpasses existing open-source and commercial methods.
\end{itemize}
\section{Related Work}
\label{sec:relatedwork}

\paragraph{Texture Generation.}
The advances of foundation models have opened new directions for automating texture generation in 3D content creation.
%
Early works leverage 2D diffusion priors via Score Distillation Sampling (SDS) to optimize 3D assets~\citep{poole2022dreamfusion,lin2023magic3d,wang2023prolificdreamer,po2024compositional,metzer2022latent,chen2023scenetex,khalid2022clipmesh,text2mesh,chen2022tango}. These approaches iteratively optimize renderings of 3D shapes to align with the distribution learned by pre-trained diffusion models~\citep{podell2023sdxl,rombach2021highresolution,esser2024sd3,lin2024common}, 
but the results are often over-saturated and unregulated for 3D shapes, making them inapplicable for practical use.
To enhance geometric fidelity, some methods~\citep{yu2023texture,bensadoun2024meta,cao2023texfusion,liu2024text} incorporate explicit 3D features, such as vertex positions, normals, or depths, to progressively inpaint the mesh across pre-defined views. Although the geometry fidelity is improved, the results are deteriorated by the synchronization process of multi-view latents.
%
%
Another line of research ~\citep{yu2023texture,zeng2024paint3d,bensadoun2024meta,yu2024texgen} projects the 3D point cloud information and supervises the training in the UV space, addressing the occlusion problem in multi-view approaches. However, they introduce topological ambiguity inherent in the UV representation, deteriorating the capability of the diffusion model to generate high-fidelity textures.

More recently, methods like MV-Adapter~\citep{huang2024mvadapter} and Hunyuan3D-Paint~\citep{zhao2025hunyuan3d} have shown promising results in generating globally consistent textures via multi-view attention~\citep{li2024era3d,kant2024spad,huang2024epidiff,shi2023mvdream,wang2023imagedream}. These methods efficiently leverage the capability of pre-trained diffusion models and the 3D geometry condition, ensuring both realistic results and spatial consistency. 

\paragraph{PBR Texture Generation.}
Recent approaches leverage pre-trained diffusion models to generate PBR materials for 3D assets.
Early works employ SDS for PBR generation and typically~\citep{chen2023fantasia3d,zhang2024dreammat,liu2024unidream,youwang2024paint,wu2023hyperdreamer,xu2023matlaber,yeh2024texturedreamer} incorporate the BRDF in the diffusion process to learn material properties.
Methods like Material-Anything~\citep{huang2024materialanything} and CLAY~\citep{zhang2024clay} iteratively denoise and synchronize latents across multiple viewpoints.
TexGaussian~\citep{xiong2024texgaussian} leverages octant-aligned 3DGS~\citep{kerbl3Dgaussians} for real-time PBR texture synthesis. Other approaches~\citep{zhang2024mapa,dang2024texpro,wang2024boosting,fang2024make} leverage LLMs to improve semantic alignment with retrieval-augmented generation. However, these methods struggle with physically grounded decomposition, view consistency, or accurate separation of albedo and MR. Some approaches~\citep{he2025materialmvp,zhu2024mcmat,vainer2024collaborative,vecchio2024matfuse,sartor2023matfusion} fine-tune image models to generate multi-channel PBR materials, but the domain gap between albedo and MR is not well addressed.

\paragraph{Image Intrinsic Decomposition.}
Material decomposition aims to estimate intrinsic materials from image inputs with unknown lighting, serving as a fundamental task in material understanding.
Early works~\citep{wimbauer2022rendering,sang2020single,boss2020two,yi2023weakly} formulate this task as an optimization problem, recovering the PBR materials from the images by solving the rendering equation~\citep{kajiya1986rendering}.
Recently, generative approaches~\citep{chen2024intrinsicanything,hong2024supermat,kocsis2024intrinsic,li2025idarb,zhu2022irisformer,huang2024blue} utilize diffusion models to decompose materials from images with promising results.
RGB-X~\citep{zeng2024rgb} and DiffusionRenderer~\citep{liang2025diffusionrenderer} propose a unified forward and inverse process for rendering and material decomposition. However, estimating individual properties from a single image with unknown illumination is ill-posed due to the inherent ambiguity between illumination and materials. Our work designs a multi-view pipeline that incorporates illumination context, enabling physically plausible and coherent material generation.

\section{Method}
\label{method}

\begin{figure*}
    \centering
    \includegraphics[width=\linewidth]{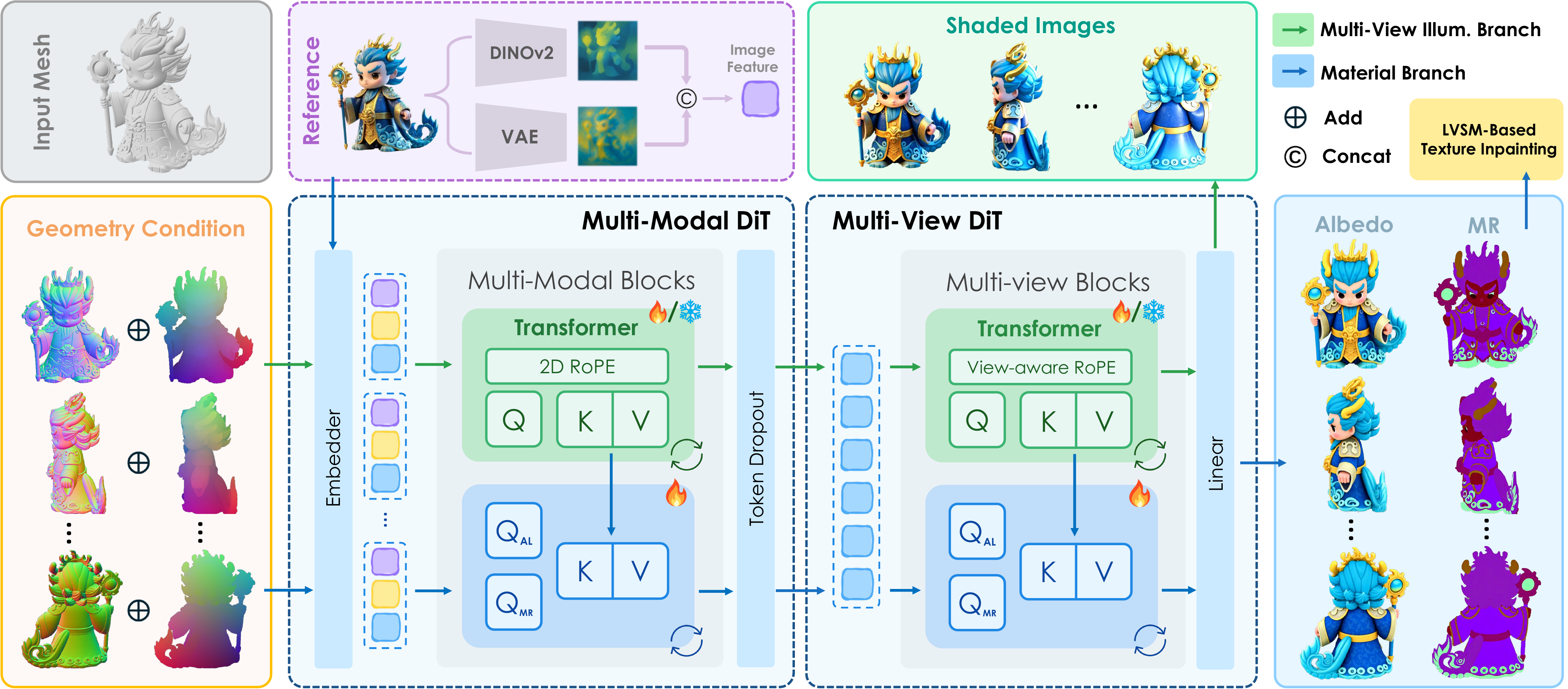}
    \caption{\textbf{Overview of \method.}  We first train a multi-view illumination-consistent base model to generate shaded images. Then, we freeze this branch and utilize its shaded features to train material branches for high-fidelity PBR texture generation. Finally, our geometry-guided LVSM synthesizes novel views from novel perspectives to enable seamless, view-consistent texture inpainting.}
    \vspace{-1.3em}
    \label{fig:method}
\end{figure*}

Given an input mesh with a reference image $\bm{I}$,  our goal is to generate $N$ view-consistent PBR images and achieve seamless material textures.
We describe the overall pipeline architecture in Section~\ref{Sec3.1}.
As shown in Fig.~\ref{fig:method}, we first train a multi-view illumination context branch that reconstructs multi-view shaded images to capture consistent lighting cues as an explicit illumination context.
This context then guides the albedo and MR branches through a lighting-aware material attention mechanism to generate multi-view material maps (Section \ref{Sec3.2}).
Finally, a geometry-guided LVSM is trained to synthesize $M$ novel views from generated $N$ views for texture inpainting (Section \ref{Sec3.3}).

\subsection{Multi-View PBR Generation Transformer}
\label{Sec3.1}
Our model integrates a multi-modal Transformer (MM-T) and a multi-view Transformer (MV-T). 
The multi-modal transformer fuses diverse modalities for each view, while the multi-view transformer enforces consistency across views.
The detailed operations are presented below.

\paragraph{Multi-Modal DiT.} The MM-T is designed to integrate geometry information, reference appearance, and material semantics (albedo or MR) for each view. Specifically, for each view $i=1,...,N$, we concatenate tokens derived from the input image encoded by VAE and DINOv2~\citep{oquab2023dinov2}, the mesh geometry (normal $\bm{N}_i$ and canonical coordinate $\bm{C}_i$) encoded by VAE, and learnable material embeddings $\bm{e}$, view latent $\latent_i$, and feed them into a stack of $l_1$ transformer blocks:
\begin{align}
\token_{\text{geo}} &=
        \text{Linear}_\text{geo}
        \big[
            \text{VAE}(\bm{N}_i) \oplus \text{VAE}(\bm{C}_i)
        \big] \in \mathbb{R}^{N \times L \times C}, \\
\token_{\text{img}} &= 
        \text{Linear}_\text{img}
        \big[
            \text{VAE}(\bm{I}) ; \operatorname{DINO}(\bm{I})
        \big] \in \mathbb{R}^{1 \times 2L \times C} , \\
\bm{z}_i^{l_1}&=\operatorname{MM-T}^{l_1}\big(
    \bm{z}_i, \token_{\text{img}}, \token_{\text{geo}}, \bm{e}
\big) \in \mathbb{R}^{N \times 3L \times C} .
\end{align}
\noindent\textbf{Multi-View DiT.}
After fusing multi-modal features in the first stage, we discard the image and domain tokens, and shift focus to enforcing cross-view consistency in the second stage.
Specifically, we concatenate their latent tokens from $N$ views into a unified sequence that is processed by a sequence of $l_2$ transformer blocks, allowing each token to interact with others for global consistency:
%
\begin{equation}
    \{\hat{\latent_i}\}_{i=1}^{N} = \operatorname{MV-T}^{l_2}(
    \bm{z}_1^{l_1}, \bm{z}_2^{l_1}, ... , \bm{z}_N^{l_1}
    ) \in \mathbb{R}^{NL \times C},
\end{equation}
where $\hat{\bm{z}}_i$ is the denoised latent for the $i$-th view, $L$ is the token length, and $C$ is the feature dimension. The whole model is trained end-to-end using a flow matching loss on multi-view images.


\subsection{Lighting-Aware Material Attention}
\label{Sec3.2}

Multi-view shaded images provide rich illumination cues for high-fidelity PBR reconstruction, serving as a reliable prior for high-fidelity PBR reconstruction. Following this insight, we design a multi-branch generative framework, where a multi-view illumination context branch provides shaded embeddings for material reasoning. The lighting-aware material branch then consumes this context to produce physically plausible PBR maps. This framework alleviates the data scarcity problem in PBR texture generation, mitigates the domain gap between albedo and MR maps, and yields more physically consistent material predictions.




\paragraph{Multi-View Illumination Context Branch.}
Unlike prior two-stage works that rely on multi-view shaded images as intermediates, we introduce a multi-view illumination context branch to learn the shaded embeddings, as illustrated in Fig.~\ref{method}. Specifically, this branch is trained to reconstruct multi-view shaded images to ensure the learned embeddings capture consistent illumination across views.
To model view-dependent illumination effects, shaded latent tokens $\bm{S} = \{\bm{s}_i\}_{i=1}^{NL}$ from all views are encoded with a view-aware RoPE~\citep{su2024roformer}, which preserves both spatial alignment and view identity. Then, we perform cross-view attention to produce shaded keys and values that condition the material branches:
\begin{equation}\label{illumattn}
\begin{gathered}
    \bm{s}_i = \sum\nolimits_j \operatorname{Softmax}_j \left( \bm{q}_i \bm{k}_j^T + \phi(t, i,j) \right) \bm{v}_j, \\
    \bm{K}_{\text{shaded}} = \bm{W}_K \bm{S}, \quad \bm{V}_{\text{shaded}} = \bm{W}_V \bm{S}.
\end{gathered}
\end{equation}
where $\phi(t,i,j)$ is a view-specific positional embedding tied to the index of query view $t$, encoding illumination relationships between views. We provide the implementation details in the supplementary. 



This design alleviates the data imbalance issue in open-source 3D datasets, where high-quality PBR maps are scarce. In this way, even data lacking plausible MR maps can be utilized, as they still provide supervision for multi-view illumination reconstruction under physically consistent materials.

\paragraph{Lighting-Aware Material Branch.}
Some works generate PBR materials via jointly modeling the albedo and MR maps, overlooking the domain gap between these two modalities.
To address this problem, we leverage the illumination context learned from the previous branch and guide albedo and MR generation separately, while conditioning both on the same shaded features to maintain the consistency of the two branches.
Specifically, we introduce a \textit{lighting-aware material attention} mechanism in which albedo and MR branches perform shaded-guided cross-attention. The shaded keys $\bm{K}_{\text{shaded}}$ and values $\bm{V}_{\text{shaded}}$ supply illumination cues to inform material-specific queries:
\begin{align}
    \text{Attn}_\text{albedo} &= \operatorname{Softmax}\left(\frac{ \bm{Q}_\text{albedo} \bm{K}_\text{shaded}^T}{\sqrt{d}}\right) \cdot \bm{V}_\text{shaded},
    \label{eq:mcaa_albedo} \\
    \text{Attn}_\text{mr} &= \operatorname{Softmax}\left(\frac{ \bm{Q}_\text{mr} \bm{K}_\text{shaded}^T}{\sqrt{d}}\right) \cdot \bm{V}_\text{shaded}.
    \label{eq:mcaa_mr}
\end{align}
This design provides a rich illumination context for the decoding process, enabling the model to separate reflectance properties from lighting effects and avoid the error accumulation problem in the two-stage design. By complementarily attending to shaded priors, the albedo branch emphasizes diffuse consistency while the MR branch captures specular characteristics, together improving both the physical plausibility and multi-view coherence of generated PBR textures.

\subsection{Texture Inpainting as Novel View Synthesis}

\label{Sec3.3}

Texture inpainting is a post-processing step that fills missing or occluded regions to ensure seamless integration with surrounding textures. Previous methods usually perform inpainting on partially textured meshes in the UV space. However, due to inherent UV discontinuities and topological ambiguities, the inpainted regions often fail to align with their surrounding areas, particularly when the UV mapping is highly fragmented, as is common for shapes produced by generative models. 

\begin{figure}[t]
    \centering
    \includegraphics[width=\linewidth]{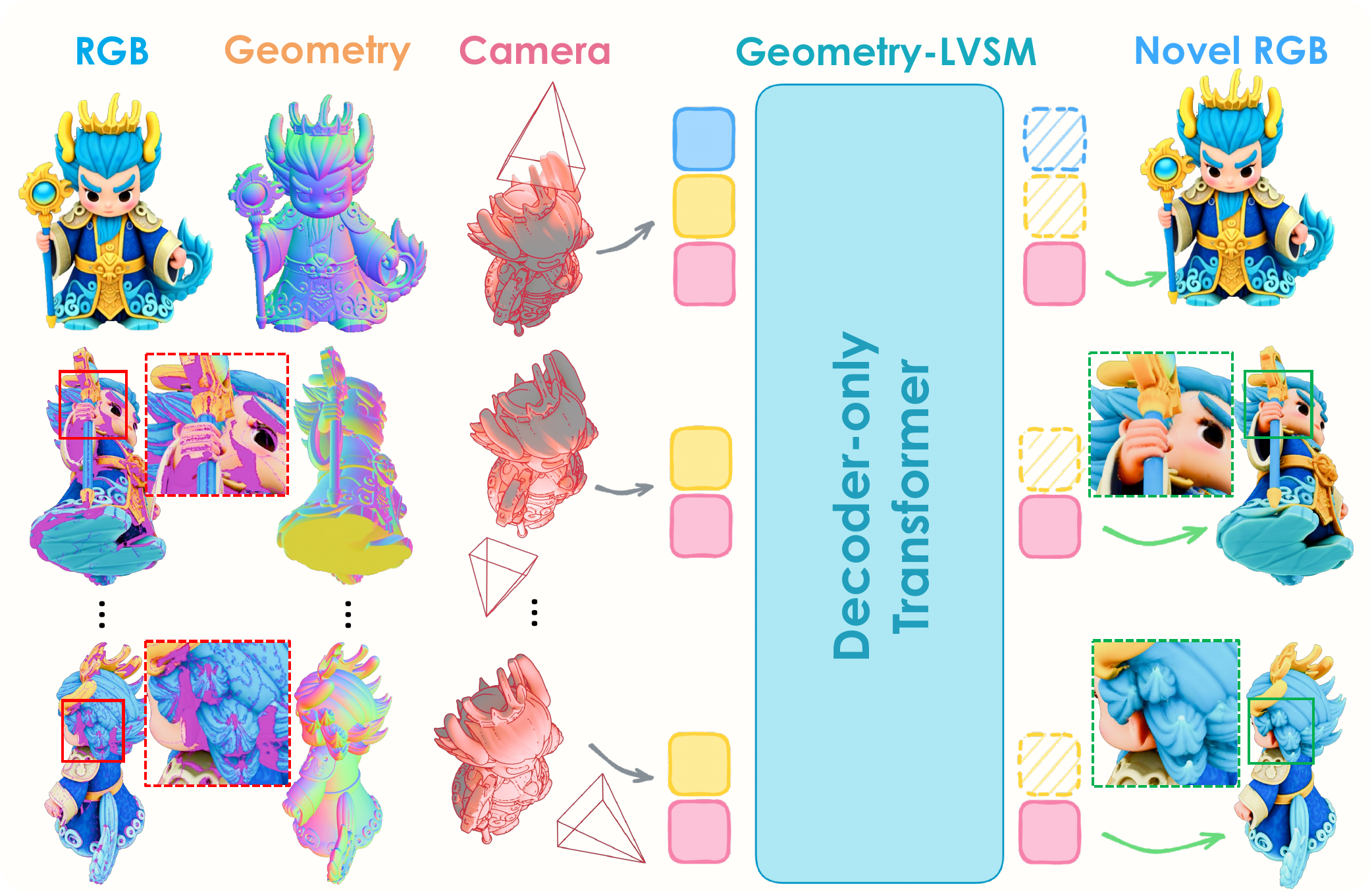}
    \caption{\textbf{Texture Inpainting with Geometry-guided LVSM.} 
    Our model infers dense novel views from sparse inputs for inpainting.}
    \vspace{-1.3em}
    \label{fig:texture_inpaint}
\end{figure}

Incomplete textures typically result from limited surface coverage in the generated sparse multi-view images. To address this, we aim to synthesize a dense set of views that fully cover the object’s surface, enabling seamless texture completion through direct back-projection.
To this end, we train a geometry-guided variant of LVSM~\citep{jin2025lvsm}, a scalable large view synthesis model known for high-quality novel view generation, to infer additional views that cover previously unobserved regions of the mesh.
These synthesized views, combined with the initial ones, form a dense set that is projected back to UV space, resulting in a seamless and complete texture.
Our framework is illustrated in Fig.~\ref{fig:texture_inpaint}.

\paragraph{Novel View Generation.} Given $N$ generated sparse view images $\{\bm{I}_i\}_{i=1}^N$, pixel-aligned Pl\"ucker ray maps $\{\bm{P}_i\}_{i=1}^N$ that encode camera intrinsics and extrinsics, and geometry conditions $\{\bm{G}_i\}_{i=1}^N$, we infer additional $M$ target views with conditions $\{\bm{P}_i^t\}_{i=1}^{M}$ and $\{\bm{G}_i^t\}_{i=1}^{M}$ to inpaint the occluded regions. To effectively embed conditions, we tokenize and map the inputs into a unified representation with a linear layer. Formally,
\begin{equation}
    \bm{x}_i = \text{MLP}([\bm{P}_i, \bm{G}_i, \bm{I}_i]),\enspace
    \bm{x}_i^t = \text{MLP}([\bm{P}_i^t, \bm{G}_i^t]) \in \mathbb{R}^d,
\end{equation}
where $d$ is the feature dimension, $\bm{x}_i$ represents the set of condition tokens, and $\bm{x}_i^t$ denotes the set of target tokens.
%
%
Following LVSM~\citep{jin2025lvsm}, we employ a decoder-only transformer to infer target views from input tokens that avoid explicit 3D representation to minimize the inductive bias:
\begin{equation}
\{\bm{y}_i\}_{i=1}^N, \{\bm{y}^t_i\}_{i=1}^M = \text{Transformer}\left(\{\bm{x}_i\}_{i=1}^N, \{\bm{x}_i^t\}_{i=1}^M\right),
\end{equation}
where the output condition tokens $\bm{y}_i$ and target tokens $\bm{y}_i^t$ are updated from the corresponding inputs.
Then, we discard condition tokens and map target tokens to the RGB space with an MLP, followed by reshaping to form the final predicted images $\hat{\bm{I}}^t$:
\begin{equation}
\hat{\bm{I}}^t_1, \dots, \hat{\bm{I}}^t_M = \text{Reshape}\left(\text{MLP}(\bm{y}^t_{1}, \dots, \bm{y}^t_{M})\right).
\end{equation}
To further improve the quality of the synthesized novel views, we adopt the test-time training method proposed in \cite{zhang2025test}.

\paragraph{View Selection.} Inspired by~\cite{zhao2025hunyuan3d}, we select target views from a predefined dense set $\mathcal{V} = \{v_1,...,v_K\}$. We first project the generated $N$ views into the UV space to obtain an incomplete texture. Then, we greedily rank the views in $\mathcal{V}$ by the area of uncovered UV regions they observe and select the top $M$ as target views.

\section{Experiments}

In this section, we evaluate our method with both open-source and commercial state-of-the-art methods, including shaded texture generation, PBR generation, and texture inpainting methods. We conduct comprehensive qualitative and quantitative comparisons, and a user study with 3D modelers, to demonstrate that our generated PBR textures align well with human perceptual preferences.

\begin{figure*}[ht]
    \centering
    \begin{minipage}{\linewidth}
        \centering
        \includegraphics[width=0.945\linewidth]{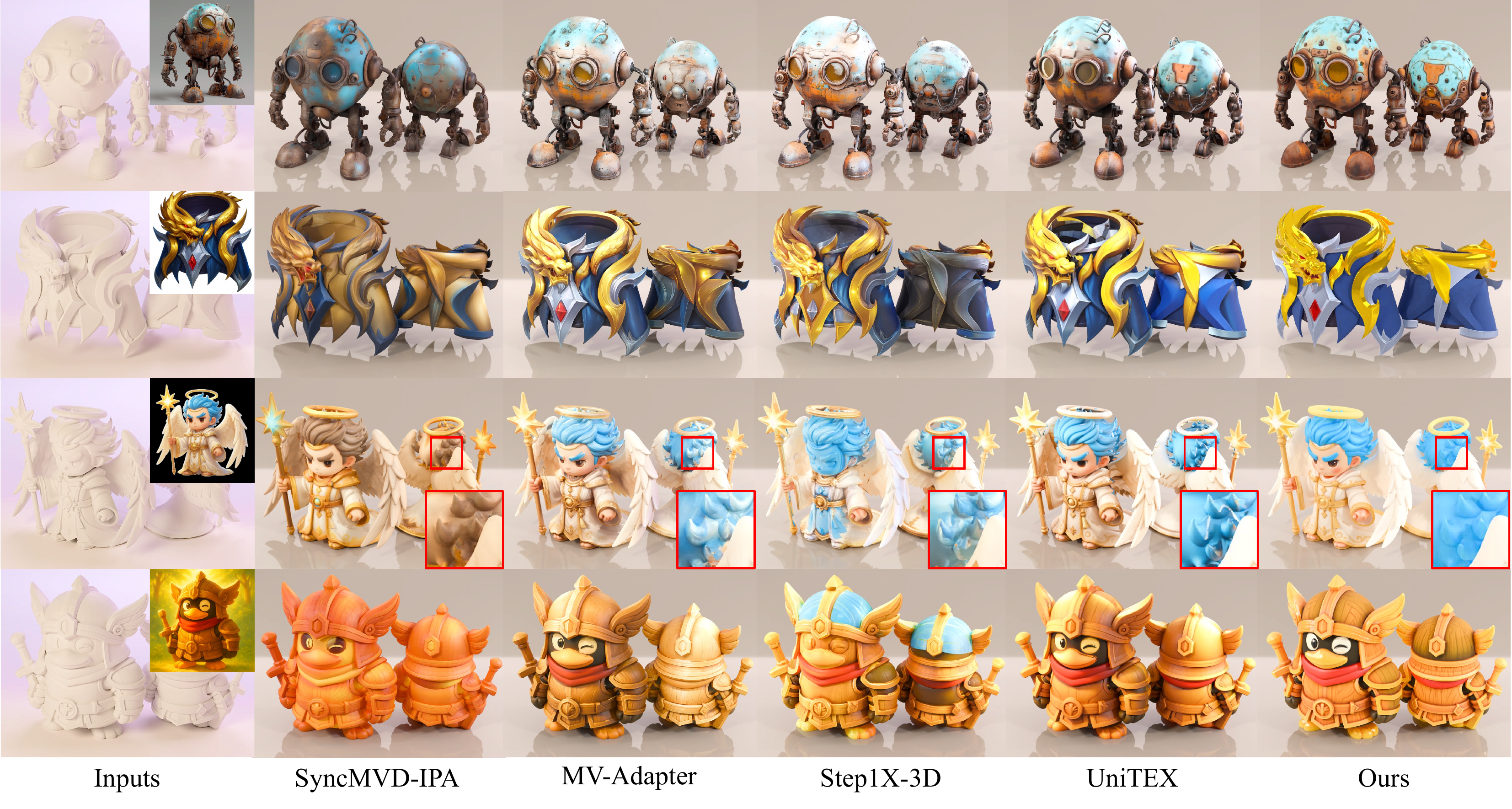}
        \vspace{-.8em}
        \caption{\textbf{Qualitative Results on Texture Generation Methods}. Our method generates plausible materials for relighting, avoids baked-in lighting, and is robust under diverse reference illuminations, including challenging cases like strong backlighting (last row).
        }
        \label{fig:comp-texture}
    \end{minipage}
    \begin{minipage}{\linewidth}
        \centering
        \includegraphics[width=0.945\linewidth]{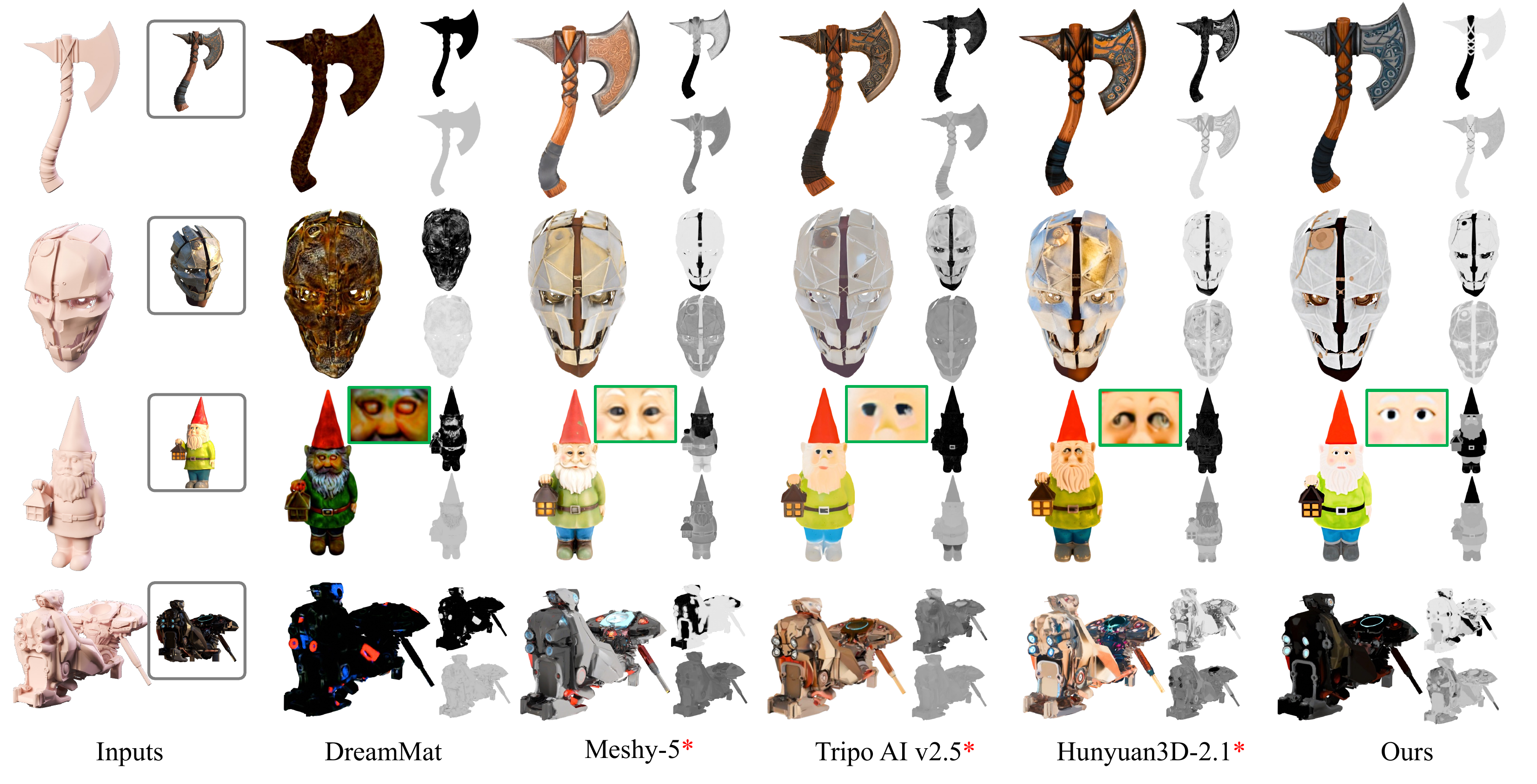}
        \vspace{-0.7em}
        \caption{\textbf{Qualitative Results on PBR Generation Methods}. Our method generates high-fidelity PBR materials, avoids light baking, and achieves competitive PBR maps compared to state-of-the-art open-source and commercial methods.
        Each object has: the albedo on the left, the metallic on the top right, and the roughness on the bottom right. \textcolor{red}{*} denotes the method trained on private datasets.}
        \label{fig:comp-pbr}
    \end{minipage}
    \vspace{-1.74em}
\end{figure*}

\subsection{Implementation Details}
\label{Sec4.1}

We curate a dataset from Objaverse and Objaverse-XL~\citep{deitke2023objaverse,deitke2023objaversexl}, containing 92K objects as our training set.
For each 3D object, we sample cameras from 30 views and render the object with 3 environment maps.
We also render multi-view albedo, metallic, roughness maps, and HDR images at a resolution of $1024\times1024$.
For the model training, we initialize our DiT from FLUX.1-dev~\citep{blackforest2024flux}, utilizing the flow matching as the training objective. We first train our shaded model on $512\times512$ and then scale up to $768 \times 768$. The entire training procedure requires approximately 106 GPU days. We provide datasets and training details in the supp.



\begin{figure*}[ht]
    \centering
    \includegraphics[width=\linewidth]{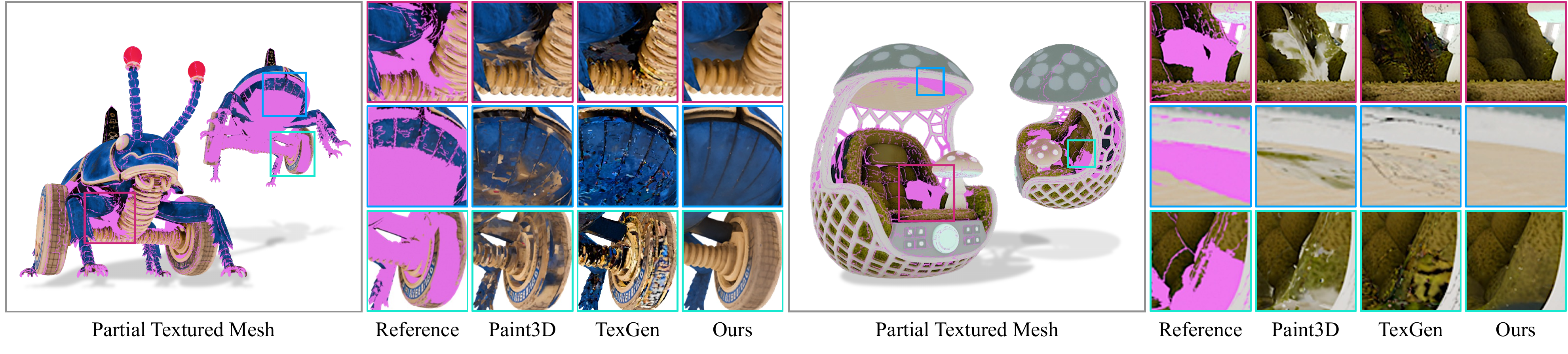}
    \vspace{-1.8em}
    \caption{\textbf{Comparison with Texture Inpainting Methods.} Our approach effectively recovers local details and exhibits greater robustness than other methods, producing semantically coherent results without visible seams.}
    \label{fig:comp-inpaint}
    \vspace{-1.2em}
\end{figure*}
\definecolor{tabfirst}{RGB}{255,204,204} 
\definecolor{tabsecond}{RGB}{255,229,204} 
\definecolor{tabthird}{RGB}{255,255,204} 

\begin{table*}[t]
\centering
\caption{Quantitative comparison with state-of-the-art methods. We compare two classes of methods, texture-only generation and PBR texture generation. Our method achieves the best performance compared with both classes. \textcolor{red}{*} denotes the method trained on private datasets.}
\setlength{\tabcolsep}{3pt}
\resizebox{0.96\textwidth}{!}{
\begin{tabular}{cc
                ccccc
                ccccc}
\toprule
& & \multicolumn{5}{c}{Texture Evaluation} & \multicolumn{5}{c}{Relighting Evaluation} \\
\cmidrule(lr){3-7} \cmidrule(lr){8-12}
Method & Type 
& {FID$\downarrow$} & {CLIP-FID$\downarrow$}  & {CMMD$\downarrow$} & {CLIP-I$\uparrow$} & {LPIPS$\downarrow$}  
& {FID$\downarrow$} & {CLIP-FID$\downarrow$}  & {CMMD$\downarrow$} & {CLIP-I$\uparrow$} & {LPIPS$\downarrow$}  \\ \midrule

SyncMVD-IPA~\citep{liu2024text}  & Texture  
&  \cellcolor{tabthird}222.1 &  \cellcolor{tabthird}21.10 & \cellcolor{tabthird}1.8263 & \cellcolor{tabthird}0.9187  & 0.2504 
&  149.1  &  18.04  & 0.7394 & 0.9101  & 0.1202 \\

MV-Adapter~\citep{huang2024mvadapter} & Texture        
&  237.3 &  24.95 &  2.4510 &  0.9022  &  0.2574    
&  123.2   &  13.82   &  0.5405  & 0.9246   &   0.1034   \\

Step1X-3D~\citep{li2025step1x}   & Texture 
&  240.9  &  24.32  &  2.2090  &  0.9053   &  0.2540
&  \cellcolor{tabthird}120.0   &   \cellcolor{tabthird}12.90  &  \cellcolor{tabthird}0.4038  &    \cellcolor{tabthird}0.9288   &  0.1000 \\

UniTEX~\citep{liang2025UnitTEX}   & Texture 
&  230.7 &  22.20  & 1.9891  &  0.9133  &  \cellcolor{tabthird}0.2473 
&  124.8   &   13.50  &  0.4707  &  0.9282  & \cellcolor{tabthird}0.0974 \\

\hline
Paint-it~\citep{youwang2024paintit} & PBR   
& 293.3  &  35.50  &  3.4137 &  0.8648  &  0.3769 
&  162.9  &  26.77   &  1.2514  &    0.8666    &   0.1564     \\

DreamMat~\citep{zhang2024dreammat} & PBR   
&  231.6 &  25.49  &  2.1722 &  0.9016  & 0.2816 
&  160.1   &  19.97  &  0.8386  &  0.8983  &  0.1346 \\

Hunyuan3D-2.1\textsuperscript{\textcolor{red}{*}}\citep{hunyuan3d2025hunyuan3d2.1}  &  PBR
& \cellcolor{tabsecond}196.6  &  \cellcolor{tabsecond}18.84  &  \cellcolor{tabsecond}1.7195  &  \cellcolor{tabsecond}0.9268  & \cellcolor{tabsecond}0.2413 
&  \cellcolor{tabsecond}103.7   &   \cellcolor{tabsecond}10.89  &   \cellcolor{tabsecond}0.3610  &   \cellcolor{tabsecond}0.9420  &  \cellcolor{tabfirst}\bfseries{0.0808}   \\

Ours & PBR         
& \cellcolor{tabfirst}\bfseries{160.8}  &  \cellcolor{tabfirst}\bfseries{14.89}  & \cellcolor{tabfirst}\bfseries{1.3669}  &  \cellcolor{tabfirst}\bfseries{0.9417}  & \cellcolor{tabfirst}\bfseries{0.1903} 
&  \cellcolor{tabfirst}\bfseries{99.6}   &  \cellcolor{tabfirst}\bfseries{10.63}  &   \cellcolor{tabfirst}\bfseries{0.3151}  &  \cellcolor{tabfirst}\bfseries{0.9436}   &   \cellcolor{tabsecond}0.0831   \\ 

\bottomrule
\end{tabular}
}
\label{tab:comparisons}
\vspace{-1.1em}
\end{table*}

\begin{figure}[ht]
    \centering
    \includegraphics[width=\linewidth]{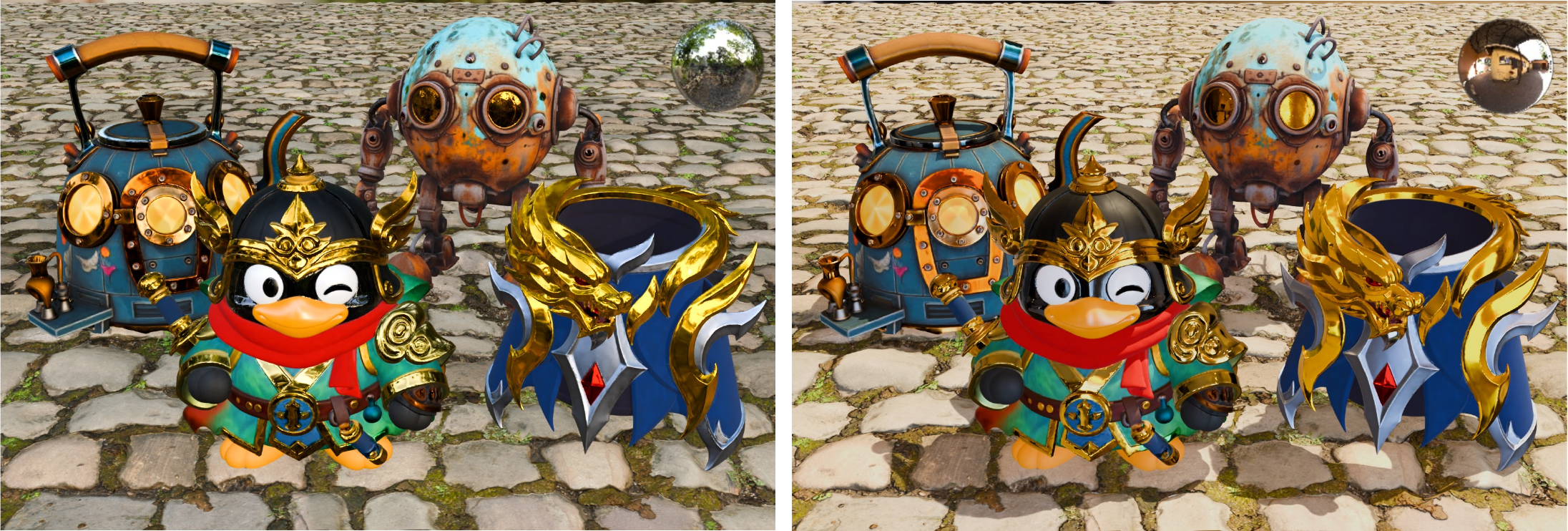}
    \vspace{-1.4em}
    \caption{\textbf{Relighting Results.} We relight generated assets under diverse lighting to show the physical plausibility of our PBR textures.
    }
    \vspace{-1.8em}
    \label{fig:relight}
\end{figure}

\subsection{Comparison with Existing Methods}

\paragraph{Baselines.}
We compare our method with comprehensive texture synthesis baselines to demonstrate its effectiveness. The baselines include shaded texture synthesis: SyncMVD~\citep{liu2024text}, MV-Adapter~\citep{huang2024mvadapter}, Step1X-3D~\citep{li2025step1x}, UniTEX~\citep{liang2025UnitTEX}; and PBR texture synthesis: Paint-it~\citep{youwang2024paintit}, DreamMat~\citep{zhang2024dreammat}, and Hunyuan3D-2.1~\citep{hunyuan3d2025hunyuan3d2.1} (MaterialMVP~\citep{materialanything} enhanced with purchased dataset).
Additionally, we compare \method~with proprietary methods such as Meshy-5~\citep{MESHY_AI} and Tripo AI v2.5~\citep{TRIPO_AI}.
We improve the original text-conditioned SyncMVD~\citep{liu2024text} by incorporating the SDXL-base model~\citep{podell2023sdxl} (the vital component) and an IP-Adapter~\citep{ye2023ip} to align with an image-to-texture task (referred to as SyncMVD-IPA).
The input images and meshes include both real-world and AI-generated examples.
For texture inpainting, we compare our method with Paint3D~\citep{zeng2024paint3d} and TexGen~\citep{yu2024texgen}.

\paragraph{Evaluation Metrics.}
We follow Hunyuan3D~\citep{hunyuan3d2025hunyuan3d2.1,lai2025hunyuan3d25highfidelity3d,zhao2025hunyuan3d} to use FID~\citep{heusel2017fid}, CLIP-FID, and LPIPS~\citep{zhang2018lpips} to evaluate texture fidelity. CLIP Maximum-Mean Discrepancy (CMMD)~\citep{jayasumana2024cmmd} assesses the diversity of the generated texture details, and CLIP-I~\citep{radford2021learning} measures prompt alignment. We further assess relighting quality for PBR materials.

\subsection{Qualitative Results}

\begin{figure*}[t]
    \centering
    \includegraphics[width=0.982\linewidth]{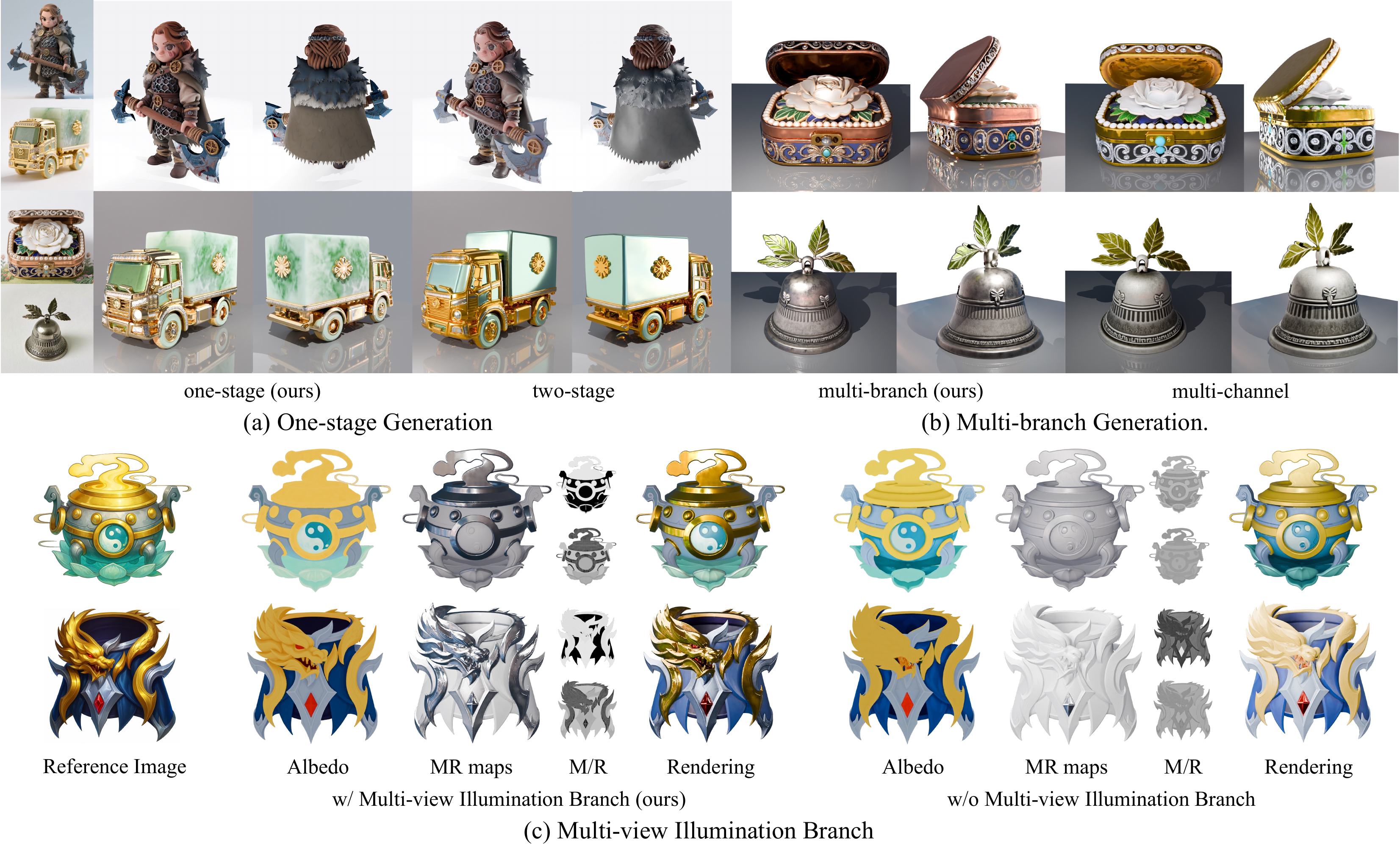}
    \vspace{-0.8em}
    \caption{
    \textbf{Ablation Study on Generation Schemes and Core Components.}
    We ablate the one-stage generation, multi-branch generation, and multi-view illumination context branch (top: metallic, down: roughness) to validate the effectiveness of our method.
    }
    \vspace{-1.6em}
    \label{fig:ablation}
\end{figure*}


We qualitatively compare LumiTex with recent baselines in texture generation and inpainting. LumiTex surpasses baselines in physical realism, lighting decoupling, and prompt fidelity.
In terms of realism, LumiTex exhibits realistic appearances under novel lighting, while others often copy input highlights as in Fig.~\ref{fig:comp-texture}, failing to produce realistic reflections due to the absence of PBR materials.
As shown in Figs.~\ref{fig:comp-texture} and \ref{fig:comp-pbr}, LumiTex effectively decouples lighting effects to ensure that models interact correctly with environment-dependent illumination. In contrast, baselines often produce diffuse maps with baked-in lighting, which deteriorates realism and hinders 3D asset reuse.
Compared to SyncMVD, Step1X-3D, and DreamMat, which exhibit color shifts, our method produces semantically faithful PBR textures that preserve fine details.
LumiTex can also generate high-quality textures across complex shapes and diverse object categories (see Fig.~\ref{fig:teaser}), retaining realistic appearances under diverse lighting (Fig.~\ref{fig:relight}), supporting downstream applications like IP design.


In Fig.~\ref{fig:comp-inpaint}, we compare texture inpainting results across baselines. Paint3D completes textures with a UV refinement model but often produces over-smoothed and semantically inconsistent results, as spatially unrelated regions are treated equally in the UV domain.
For example, the Beetle's underside and wheels are inpainted without spatial awareness, leading to texture bleeding.
TexGen mitigates this issue via 3D spatial encoding, yet still suffers from disrupted semantic coherence and seams, due to the inherent topological ambiguities of UV mapping.
Our method, conditioned on 2D views and 3D cameras, avoids the adverse effects caused by UV representation. By unifying condition and target views in a framework, our method preserves fine details while ensuring global consistency.


\subsection{Quantitative Results}

We evaluate 133 objects excluded from training on two aspects: texture quality and relighting evaluation. Texture quality assesses fidelity and alignment with the reference image, while relighting evaluation involves rendering each object from 32 views sampled on the Fibonacci sphere with random environment maps and comparing them against ground-truth renderings.
As shown in Tab.~\ref{tab:comparisons}, our method surpasses both texture-only and PBR-based baselines, achieving better FID and LPIPS for visual fidelity, lower CMMD for richer and more diverse texture generation, and stronger semantic alignment with the reference images as reflected by higher CLIP-I and lower CLIP-FID scores.


\definecolor{tabfirst}{RGB}{255,204,204}  
\definecolor{tabsecond}{RGB}{255,229,204} 
\definecolor{tabthird}{RGB}{255,255,204}  

\begin{table}[t]
\centering
\caption{
User study results on overall quality, texture completeness, and PBR material qualities.
}
\setlength{\tabcolsep}{2pt}
\resizebox{0.98\linewidth}{!}{
\begin{tabular}{lccccc}
\toprule
Method & Quality$\uparrow$ & Compl.$\uparrow$ & Diffuse$\uparrow$ & Metallic$\uparrow$ & Rough.$\uparrow$ \\
\midrule
SyncMVD~\citep{liu2024text}       & 2.29 & \cellcolor{tabthird}3.27 & 2.39 & --   & --   \\
MV-Adapter~\citep{huang2024mvadapter}    & 2.65 & 3.06 & 2.65 & --   & --   \\
Step1X-3D~\citep{li2025step1x}     & 2.70 & 2.98 & 2.58 & --   & --   \\
UniTEX~\citep{liang2025UnitTEX}        & \cellcolor{tabthird}2.96 & 2.98 & \cellcolor{tabthird}2.75 & --   & --   \\
Paint-it~\citep{youwang2024paintit}     & 2.27 & 2.97 & 2.19 & \cellcolor{tabthird}2.46 & \cellcolor{tabthird}2.57 \\
DreamMat~\citep{zhang2024dreammat}      & 2.05 & 2.92 & 2.58 & 2.40 & 2.33 \\
Hunyuan3D-2.1\textsuperscript{\textcolor{red}{*}}\citep{hunyuan3d2025hunyuan3d2.1} 
& \cellcolor{tabsecond}3.69 & \cellcolor{tabsecond}3.98 & \cellcolor{tabsecond}3.57 & \cellcolor{tabsecond}3.34 & \cellcolor{tabsecond}3.61 \\
Ours          
& \cellcolor{tabfirst}\bfseries 4.48 
& \cellcolor{tabfirst}\bfseries 4.61 
& \cellcolor{tabfirst}\bfseries 4.34 
& \cellcolor{tabfirst}\bfseries 4.14 
& \cellcolor{tabfirst}\bfseries 4.07 \\
\bottomrule
\end{tabular}}
\label{tab:userstudy}
\vspace{-1.4em}
\end{table}

To evaluate perceptual quality, we conduct a user study, where 23 3D modelers rate results (ranging from 1 to 5) generated from different methods on five criteria that are difficult to quantify: rendering quality, albedo, roughness, metallic accuracy, and texture completeness,  with input images and meshes provided for reference.
As shown in Tab.~\ref{tab:userstudy}, our method outperforms all baselines across all criteria, demonstrating strong alignment with human preference.

\subsection{Ablation Study}

\noindent\textbf{One-Stage Generation.}
To validate the effectiveness of our end-to-end pipline, we compare it to a two-stage variant that first generates multi-view shaded images, followed by a PBR decomposition model. This variant is fine-tuned from our illumination context branch and IDArb~\citep{li2025idarb}.
As shown in Fig.~\ref{fig:ablation}(a), it often produces inaccurate material predictions, such as excessive metallicity, plastic-like surfaces, or overly uniform albedos, likely due to error accumulation across stages.
In contrast, our unified design produces more realistic results, showing stronger fidelity for this task.

\noindent\textbf{Multi-Branch Generation.}
To evaluate our disentangled multi-branch design, we compare it with a multi-channel variant that jointly predicts albedo and MR in a unified output space, following prior works~\citep{he2025materialmvp, zhang2024clay}. As shown in Fig.~\ref{fig:ablation}(b), this joint prediction often results in inaccurate outputs, especially in metallic regions. Our illumination-guided, material-specific design maintains channel separation and produces more physically and semantically accurate results.

\noindent\textbf{Multi-View Illumination Context Branch.}
We show the importance of illumination context branch by training a material-only variant on the same dataset until convergence. As shown in Fig.~\ref{fig:ablation}(c), the variant fails to produce accurate MR maps in metallic regions, resulting in a plastic-like appearance under relighting.
The illumination context contributes greatly to accurate PBR generation.


\section{Conclusion}

We present LumiTex, an end-to-end multi-branch pipeline for high-fidelity PBR texture generation.  By combining a multi-view illumination context branch with a novel lighting-aware material attention mechanism, LumiTex enables physically plausible PBR map generation. To ensure global surface coverage and coherence, we train a geometry-guided LVSM for texture inpainting.
Extensive experiments demonstrate that LumiTex outperforms existing methods in texture quality, semantic alignment, and relighting fidelity, offering a practical solution for PBR texture generation.

\newpage
{
    \small
    \bibliographystyle{ieeenat_fullname}
    \bibliography{main}
}
\newpage
\appendix


\end{document}